# An Agentic System for Schema Aware NL2SQL Generation

David Onyango[1], Naseef Mansoor[2]
[1] *Project Path, Boston, MA, USA* [2] *Minnesota State University Mankato, MN, USA*
david.onyango@projectpath.ai , naseef.mansoor@mnsu.edu

*Abstract*— The natural language to SQL (NL2SQL) task plays a pivotal role in democratizing data access by enabling non-expert users to interact with relational databases through natural language. While recent frameworks have enhanced translation accuracy via task specialization, their reliance on Large Language Models (LLMs) raises significant concerns regarding computational overhead, data privacy, and real-world deployability in resource-constrained environments. To address these challenges, we propose a schema based agentic system that strategically employs Small Language Models (SLMs) as primary agents, complemented by a selective LLM fallback mechanism. As the LLM is invoked only upon detection of errors in SLM-generated output, the proposed system significantly minimizes computational expenditure. Experimental results on the BIRD benchmark demonstrate that our system achieves an execution accuracy of 47.78% and a validation efficiency score of 51.05%, achieving over 90% cost reduction compared to LLM-centric baselines as approximately 67% of queries are resolved using local SLMs. The system achieves an average cost per query of $0.0085 compared to $0.094 for LLM-only systems, achieving near-zero operational costs for locally executed queries.

## I. INTRODUCTION

The growing need for data-driven decision-making has elevated the democratization of database access to a strategic priority for businesses. The Natural Language to SQL (NL2SQL) task for translating natural language questions into executable structured query language (SQL) statements has emerged as an enabler, bridging the gap between non-technical users and complex relational data systems. [1]. This transformation has progressed significantly from early rule-based systems in the 1970s to modern systems powered by Large Language Models (LLMs) [2]. Several benchmark databases like BIRD [3], Spider [4], and WikiSQL [5] has been designed to facilitate the design and evaluation of such systems. Despite recent advances, deploying NL2SQL systems in production faces three core challenges: (1) the high computational cost of large language models (LLMs); (2) data regulations mandating on-premise deployment; and (3) accurate results on complex queries involving intricate schemas, multi-table joins, and domain-specific constraints remain difficult.

Recent advances in multi-agent NL2SQL systems show that decomposing the task among specialized agents significantly boosts performance [6]. MAC-SQL [7], a pioneering method, achieves 59.59% execution accuracy on the BIRD benchmark by distributing tasks across collaborative agents that emulate human problem-solving. However, it relies heavily on costly LLMs, undermining its scalability. Current systems also face critical limitations: handling complex enterprise schemas with hundreds of tables, resolving ambiguity from colloquialisms and domain jargon, ensuring robustness under distribution shifts, and effectively integrating execution feedback.

In this paper, we introduce a novel agentic system that synergistically integrates Small Language Models (SLMs) as the primary inference engine with LLMs deployed selectively as an intelligent fallback mechanism. The proposed system decomposes the NL2SQL task into a coordinated pipeline comprising four specialized modules: schema extraction, query decomposition, SQL generation, and query validation. Each module is handled by an agent. To mitigate hallucination and enforce strict adherence to database constraints, the proposed system incorporates schema-aware prompting and iterative refinement loops that progressively align generated queries with the underlying schema semantics. We benchmark our system against LLM-based multi agent systems on the BIRD dataset. Our system achieves an execution accuracy of 47.78% and a validation efficiency score of 51.05%, while demonstrating superior cost efficiency relative to purely LLM-driven baselines. Notably, on the BIRD benchmark, approximately 67% of queries are successfully resolved using only the SLM, substantially reducing inference costs and enabling on-premises deployment. The key contribution of this work include:

1. Design of an agentic system for NL2SQL task with SLM as a primary inference engine and intelligent LLM fallback for error recovery.
2. A comparative study of the proposed system against state-of-the-art LLM-based NL2SQL approaches, evaluating performance along three key dimensions: execution accuracy, validation efficiency score, and cost per query.

The remainder of this paper is structured as follows: Section II covers related works; Section III details the architecture of the schema-aware agentic NL2SQL system; Section IV presents the experimental results; Section V discusses the study's limitations; Section VI provides concluding remarks; and Section VII discuss future directions.

## II. RELATED WORKS

Over the past five decades, NL2SQL systems have undergone significant evolution across multiple technological stages. The earliest approaches, exemplified by the LUNAR system introduced in 1973, primarily depended on manually crafted linguistic rules and domain-specific patterns. While these systems achieved high accuracy within constrained application domains, their reliance on rigid, predefined patterns severely limited adaptability and scalability for broader, open-domain database querying tasks [8]. The advent of deep learning marked a transformative shift in NL2SQL research, introducing sequence-to-sequence architectures capable of learning intricate mappings between natural language queries and SQL statements without the need for explicit rule engineering [9] [10]. The introduction of the Spider benchmark marked a pivotal advancement in NL2SQL research. By establishing a comprehensive and standardized evaluation framework for cross-domain querying, this benchmark facilitated the rapid progress in the field, driving the development of increasingly sophisticated neural architectures capable of generalizing across diverse database schemas and natural language variations [4]. Transformer architectures pushed this improvement further, especially with pretrained models, such as BERT and T5 [11].

The emergence of Large Language Models (LLMs) has marked a transformative shift in NL2SQL research. LLM based systems have shown strong proficiency in complex subtasks such as schema understanding, enabling more flexible, context-aware, and generalizable interpretations of natural language queries compared to earlier rule-based or domain-specific approaches [12] [13]. However, despite their remarkable capabilities, the adoption of LLMs in NL2SQL systems raises several challenges, particularly due to their substantial computational demands, which drive up operational costs and hinder scalability. Moreover, concerns surrounding data privacy and security; especially utilization of proprietary LLMs poses additional barriers to widespread implementation in enterprise and sensitive database environments [14] [15]. Despite their robust capabilities, LLMs demand substantial computational resources and processing power, which poses challenges for widespread adoption. This high resource requirement limits accessibility for smaller organizations and constrains their deployment in privacy-sensitive contexts where data confidentiality and local processing are critical [16]. Furthermore, these proprietary models raise concerns regarding data privacy, as sensitive information may be exposed during processing [17].

Recent studies highlight that while modern LLM-based NL2SQL models effectively capture question semantics, they often struggle to align generated SQL queries with database schemas—particularly in handling primary and foreign keys. Such misalignment commonly results in errors including incorrect table joins, misuse of grouping columns, inappropriate logical connectors (e.g., AND vs. INTERSECT), and mismatched entity values (e.g., "USA" vs. "United States of America") [18]. Building on the earlier approaches, SA-SQL introduced a three-component draft-and-correct framework [18]. The framework uses a selector to filter irrelevant tables and columns, a generator to produce an initial SQL query from the selected schema, and a corrector to fix alignment errors, improving both efficiency and accuracy.

Building on this evolution, multi-agent collaboration in NL2SQL marks a departure from monolithic designs. The MAC-SQL framework is a notable example, utilizing three specialized agents to handle different aspects of query generation [7]. Building on query decomposition strategies, DAIL-SQL introduced a demonstration-driven approach that breaks queries into sub-problems and uses semantically similar examples retrieved via dense embeddings to guide LLM-based SQL generation with few-shot prompting [19]. DIN-SQL further enhanced this approach by incorporating iterative self-correction, using execution feedback and error diagnostics to refine queries across multiple cycles, achieving strong performance across varied domains and query complexities [20].

Alternatively, SLMs have emerged as an attractive alternative to LLMs due to their efficiency and low-cost operation. Such models can be also applied to NL2SQL tasks. By breaking complex queries into simpler components and leveraging targeted fine-tuning, SLMs can narrow the performance gap with larger, proprietary models [17]. Open-source initiatives like CodeS have shown that strategic pre-training and bi-directional data augmentation can enhance SQL generation, improving both accuracy and robustness in challenges such as schema linking and domain adaptation, highlighting the versatility of SLMs across diverse application domains [21]. In this paper, we explore how the integration of SLMs into NL2SQL systems. By dividing the NL2SQL generation task into multiple specialized agents utilizing SLMs, we propose a new system that is extremely cost efficient compared to their LLM based counterparts.

III. METHODOLOGY

In this section, we present the architecture of the proposed schema aware agentic NL2SQL system. Instead of relying on a single, monolithic model for query generation, the system divides the process into specialized subtasks: schema extraction; decomposition; query generation; execution and validation. Each of the tasks is handled by a dedicated specialized agent. The overall system architecture is illustrated in Fig. 1, followed by a detailed explanation of each agent's role and function in the following subsections.

*A. The Extractor Agent*

The extraction agent is the core of our system, designed to collect and synthesize relevant database contexts. Traditional methods often fail to scale with large enterprise databases containing hundreds of tables and complex relationships. To overcome this, our schema-aware mechanism uses a multi-source retrieval strategy that combines three key information channels: (1) database metadata describing structure and constraints, (2) Retrieval-Augmented Generation (RAG) contexts that extract semantic details from documentation such as table meanings, column definitions, and business rules, and (3) evidence mappings from historical query-schema alignments capturing domain-specific language patterns. This combined approach enables a deeper understanding of complex database schemas and minimizes model hallucination.

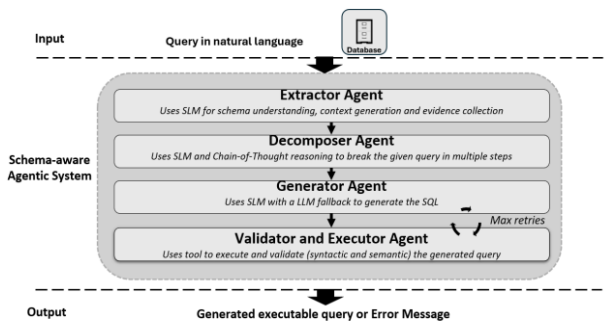

**Fig. 1.** Architecture of the Schema Based Agentic System for NL2SQL

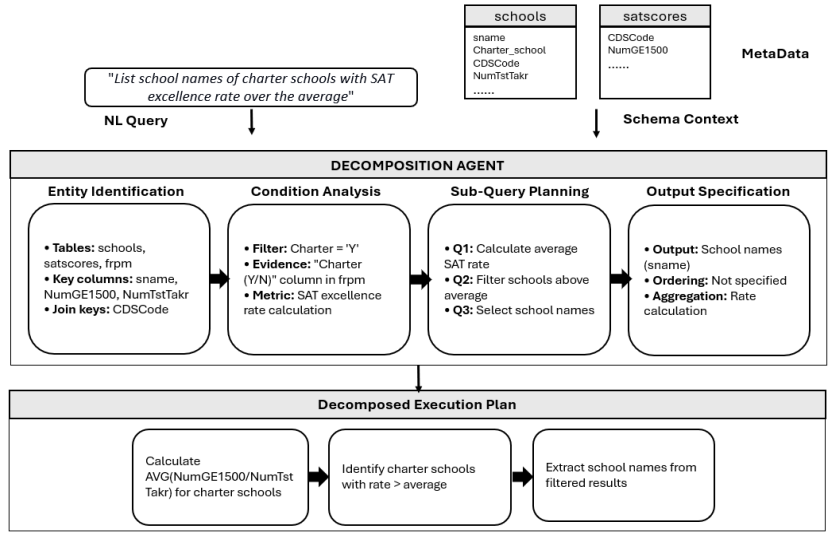

**Fig. 2.** Generation of decomposition plan by the Decomposer Agent.

The system uses vector embeddings to map both user queries and database documentation into the same semantic space. It performs an approximate nearest-neighbor search to retrieve the top 10 most relevant documentation segments. These are then re-ranked and filtered to highlight key contexts such as table links, column meanings, and domain constraints. By using precomputed embeddings, retrieval time drops from 5–10 seconds to under one second per query. Combining structural schema data with contextual evidence reduces hallucination, improves constraint handling, and strengthens entity and relationship recognition in complex, multi-table databases. The resulting schema-aware context is then passed to the next stage in the pipeline as shown in Fig. 1.

*B. The Decomposer Agent*

The decomposer agent breaks complex natural language questions into smaller, easier-to-handle steps using guided reasoning. It follows a four-stage process: first, it identifies key entities by linking words in the query to elements in the database; second, it analyzes conditions to extract filters and constraints; third, it plans how different parts of the query should be executed and in what order; and finally, it determines what information should appear in the results and how it should be formatted. The agent uses a schema-aware context provided by the extractor agent containing information on the database structure, data dictionary, and supporting evidence. This allows the decomposer to accurately match natural language terms with the right database components. This structured approach is inspired by hierarchical task planning that allows the system to handle complex queries involving multiple steps, joins, nested operations, and references. The resulting intermediate representation serves as a clear, structured blueprint for later SQL generation, making the overall process faster and more efficient. The workflow of the decomposed agent is demonstrated in Fig. 2.

As an example, consider the query "List school names of charter schools with SAT excellence rate over the average". To process this, the agent first identifies key entities and matches them to database elements. For instance, "charter schools" maps to schools where the charter flag is marked as "Y," and "SAT excellence rate" refers to a calculated value that combines data from both the schools and satscores tables. This mapping step is crucial because "SAT excellence rate" is not a direct column in the database, and it must be computed from existing attributes. Next, the agent interprets the phrase "over the average" as a comparison that requires a nested subquery to determine the overall average SAT excellence rate before filtering results. During planning, the agent ensures that the average is computed first, then used to identify schools exceeding that value. Finally, the system specifies what to display. In this case, school names are shown and may include optional sorting in alphabetical order. The result is a structured execution plan outlining entities, conditions, subqueries, and outputs, which later guides the automatic generation of the corresponding SQL query.

*C. The Generator Agent*

Provided with the decomposition plan, database context, and evidence mapping, the generator agent is responsible for creating a valid SQL query from a natural language input. To achieve this, it uses a multi-stage generation process with models designed for different situations. In the first stage, a smaller language model (Llama-3.1, 8B) generates an initial SQL query. This query is then sent to the executor and validator agent for testing. If the query fails to run, a fallback mechanism activates a LLM (GPT-4o) to regenerate the SQL. This model uses the original query, the failed SQL, and the error message to make precise corrections. The system allows up to three regeneration attempts before reporting a generation failure to the user. The overall query generation process is illustrated in Figure 3.

### D. The Validator and Executor Agent

The Validator/Executor Agent ensures query correctness through a four-stage validation process, enhanced with evidence-based autocorrection. Before execution, the agent performs evidence-based value validation by comparing SQL predicates against evidence mappings extracted during schema retrieval. This auto-correction mechanism identifies and fixes common value mismatches, such as using natural language terms (e.g., "East Bohemia") instead of exact database values (e.g., "east Bohemia") as documented in evidence mappings. The syntax validation ensures the accuracy of SQL syntax, verifies the correct formation of JOIN clauses, checks column and table name references against the schema, and confirms compatibility with the database. The execution validation executes the query within the actual database in a secure transaction environment, identifies runtime errors such as missing columns, type mismatches, and constraint violations, and records execution time for performance evaluation. Lastly, the semantic validation assesses the structure of the result set to ensure it matches the expected output format, confirms that aggregations and groupings are consistent with the original query's intent, and identifies empty results that could suggest logical errors. This multi-stage validation framework goes beyond simple syntax checking by ensuring both syntactic and semantic correctness to the user intent. Once the generated query is validated, it is executed by the executor to verify the correctness of the generated query. During the execution step, if an error is encountered (e.g., due to syntax, execution failures, and/or constraint violations), the executor passes detailed diagnostic information back to the generator for triggering the LLM fallback. The error diagnostics include: (1) execution error messages, (2) validation warnings flagged during pre-execution checks, and (3) the original failed SQL for comparison. After three retries of the LLM fallback mechanism to generate a valid query, this agent will report an error. Otherwise, the resultant query is shown to the user along with the results.

## IV. EXPERIMENTAL RESULTS

In this section, we discuss the implementation of the proposed agentic system and the experimental results for the proposed system when compared to LLMs only agentic systems.

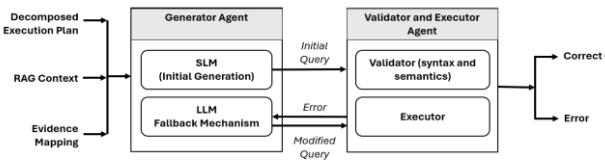

**Fig. 3.** Collaboration between the Generator Agent and the Validator and Executor Agent for generating output from decomposed execution plan, RAG context, and evidence mapping.

### A. Implementation of the proposed agentic system

The agentic system is implemented using LangGraph [22] as the workflow orchestration framework, with LangChain components managing inter-agent communication, stateful memory, and context routing. The extraction agent uses a compact embedding model (all-MiniLM-L16-v2) to convert both user queries and database documentation into numerical representations, which are stored in Chroma DB for quick and efficient retrieval of related information. The decomposition agent applies the Mistral-7B SLM to break down natural language questions into structured steps, using schema-aware context. For generating the actual SQL queries, the Llama-3.1-8B [23] model serves as the main agent. If an initial query fails validation, an intelligent fallback mechanism automatically engages GPT-4o to analyze the errors, refine the query, and retry up to three times before reporting a failure. Table 1 shows the models used in implementing the agentic system. The code is available at https://github.com/mindslab25/CESMA.

### B. Benchmark Dataset

To evaluate the proposed system, we adopted the BIRD benchmark which contains 12,751 text-to-SQL pairs, 95 databases and covers 37 professional domains, such as sports, healthcare, and politics. The databases were sourced from real-world platforms to ensure realistic value distributions and schemas. It focuses on new challenges, such as handling dirty/noisy database values, requiring external knowledge to map natural language to database values, and the need for efficient SQL generation for large databases. For this study, we used the BIRD development set with 1534 text-to-SQL pairs on 11 databases for evaluation.

### C. Evaluation Metrics

BIRD employs two primary evaluation metrics to assess text-to-SQL model performance across accuracy and efficiency dimensions. Execution Accuracy (EX) is a metric that is used to measure the proportion of predicted SQL queries that produce the exact same result set as the ground-truth SQL queries when they are executed. Formally, if $V_i$ denote the result set obtained by executing the $i^{th}$ ground-truth SQL query $Y_i$, and the result is $\hat{V}_i$ denote the result set obtained by executing the corresponding predicted SQL query $\hat{Y}_i$ then EX metric is computed as follows:

$$EX = \frac{1}{|Q|} \sum_{i=1}^{|Q|} I(V_i, \hat{V}_i) \qquad (1)$$

Here, $|Q|$ is the total number of queries and I(.) represents the indicator function. This metric captures functional correctness by evaluating query execution outcomes rather than syntactic similarity, thereby accommodating semantically equivalent SQL formulations that may differ structurally.

**Table. 1.** Implementation of the proposed agentic system for NL2SQL generation.

| Task | Model Name | Agent |
|---|---|---|
| Embedding | All-MiniLM-L6-v2 | Extractor (RAG) |
| Reasoning, Code Generation (Local) | Mistral (7B) | Decomposer |
| | Llama 3.1 (8B) | Generator |
| | | Executor/Validator |
| Code Generation (Fallback) | GPT-4o | SQL Generator |

Another metric introduced by BIRD is the Valid Efficiency Score (VES). This metric is used to measure the efficiency of valid (i.e., correctly executed) SQL queries. It only considers queries that pass the 'EX' check. An incorrect query is considered useless, regardless of its speed, and receives no score. VES provides a comprehensive evaluation by combining accuracy and efficiency, which is crucial for real-world applications on large databases. Formally, VES is computed by the following equation,

$$VES = \frac{1}{|Q|} \sum_{i=1}^{|Q|} I(V_i, \hat{V}_i) \cdot R(Y_i, \hat{Y}_i) \qquad (2)$$

Here, R(.) is the relative execution efficiency of predicted SQL in comparison to ground-truth SQL computed as $\sqrt{E(Y_i)/E(\hat{Y}_i)}$. The E(.) is the absolute execution efficiency for each SQL in a given environment.

### D. Comparative Performance Analysis

For the comparative performance analysis, we evaluated our proposed model against three state-of-the-art text-to-SQL systems: MAC-SQL, DAIL-SQL, and DIN-SQL. Table. 2 summarizes the comparison of these baseline models with our system based on execution accuracy (EX) and validation efficiency score (VES). As shown in the table, all large language model (LLM)-based multi-agent systems outperform our proposed hybrid approach. MAC-SQL, a pioneering agentic system, divides the SQL generation process among specialized agents such as schema selectors, candidate generators, and refiners. Its collaborative reasoning and iterative refinement mechanisms enable it to achieve an execution accuracy of 59.59% on the BIRD benchmark. DAIL-SQL, on the other hand, uses a decomposition-based prompting strategy that breaks complex queries into simpler sub-problems. It employs semantically retrieved demonstration examples to guide LLM-based SQL generation but performs slightly worse than MAC-SQL on both EX and VES. DIN-SQL extends this paradigm by incorporating self-correction mechanisms that iteratively refine queries using execution feedback and error diagnostics. However, it records the lowest performance among the LLM-based systems, with 55.90% EX and 59.44% VES. In comparison, our proposed hybrid agentic system has an EX of 47.78% and a VES of 51.05%.

### E. Comparatice Analysis of Cost Per Query:

All three baselines are LLM-focused approaches that rely mainly on GPT-4 as their reasoning engines. In contrast, our hybrid agentic system uses LLM only as a fallback, which offers substantial cost savings, particularly when the SLM runs on local machines. Table 3 shows the cost per query for both the baseline and proposed systems. The average cost per query for the proposed hybrid system is calculated using weighted average of the SLM and LLM costs ($2.50/1M input and $10/1M output tokens). For baseline systems (MAC-SQL, DAIL-SQL, DIN-SQL), we estimate costs by applying GPT-4 standard pricing ($30/1M input tokens, $60/1M output tokens) to the observed token consumption pattern. The table highlights that our system achieves over 90% cost reduction compared to LLM-only approaches, with an average cost per query of $0.0085 compared to $0.094 for LLM-centric systems. This improvement comes from the fact that approximately 67% of queries are handled by the zero-cost local SLM, achieving near-zero operational costs for locally executed queries, with only a small portion requiring the LLM as a fallback.

## V. DISCUSSION AND LIMITATIONS

The experimental results illustrate a critical trade-off between computational efficiency and execution accuracy in NL2SQL systems. Although, due to the deep reasoning capabilities, LLM-based baselines like MAC-SQL achieve superior execution accuracy (59.59%), our hybrid architecture demonstrates that approximately 67% of standard queries can be successfully resolved using local SLMs. This high percentage of SLM-based queries results in over 90% cost reduction, with an average cost per query of $0.0085 compared to $0.094 for LLM-centric systems.

Despite the promising cost-efficiency balance, the current study and proposed system exhibit several limitations that must be acknowledged:

*Accuracy Gap in Complex Reasoning:* The system exhibits performance degradation on queries requiring nested subqueries, joins, or complex temporal reasoning due to the SLM based query generation. The intelligent fallback mechanism mitigates this but does not fully bridge the reasoning gap inherent to smaller language models.

*Latency Overhead in Fallback Scenarios:* While the average latency is low, the tail latency is significant due to the LLM based fallback loop. The "retry logic" results in a slower response time for the most difficult queries, impacting the user's experience in real-time applications.

**Table. 2.** Comparative performance analysis of the proposed model with MAC-SQL, DAIL-SQL, DIN-SQL.

| NL2SQL System | Language Model | Fine Tuned | Execution Accuracy (%) | Validation Efficiency Score (%) |
|---|---|---|---|---|
| MAC-SQL | GPT-4 | | 59.59 | 67.68 |
| DAIL-SQL | GPT-4 | | 57.41 | 61.95 |
| DIN-SQL | GPT-4 | | 55.90 | 59.44 |
| **Proposed Agentic** | Mistral, Llama, GPT-4o | No | 47.78 | 51.05 |

**Table. 3.** Comparative cost per query of the proposed model, MAC-SQL, DAIL-SQL, DIN-SQL for BIRD benchmark.

| NL2SQL method | Primary Model | Cost Per Query |
|---|---|---|
| MAC-SQL | GPT-4 | ~0.094 |
| DAIL-SQL | GPT-4 | ~0.094 |
| DIN-SQL | GPT-4 | ~0.094 |
| Agentic System (without Fallback) | Llama | ~0.00 |
| Agentic System (with Fallback) | GPT-4o | ~0.0015 |

*Cascading Error Propagation:* Due to the sequential dependency between agents, errors generated by earlier agent will impact the overall outcome. For example, if the decomposer agent generates incorrect entity linking or flawed planning, it will impact the overall outcome even though downstream agents may have correctly performed their tasks.

*Dependency on Schema Naming Conventions:* The extractor agent relies heavily on semantic similarity between user queries and database schema documentation. In legacy databases with cryptic column names and sparse documentation, the retrieval accuracy drops, leading to increased hallucination.

## VI. CONCLUSION

In this work, we introduced a hybrid agentic system for the NL2SQL task. Our hybrid design leverages SLM as primary agents for initial query generation, with LLMs serving as intelligent fallback mechanisms. The proposed system is able to achieve competitive execution accuracy at a fraction of the cost. On the BIRD benchmark, the system reaches 47.78% execution accuracy and 51.05% validation efficiency, achieving over 90% cost reduction compared to LLM-centric baselines as approximately 67% of queries are handled entirely by local SLMs. The system achieves an average cost per query of $0.0085 compared to $0.094 for LLM-only systems, incurring near-zero operational expenses for locally executed queries. Additionally, the proposed system facilitates data privacy and supports on-premises deployment, which ensures compliance with strict regulatory requirements. This makes our proposed system a secure, efficient, and cost-effective alternative to fully LLM-based solutions for real-world organizational applications.

## VII. FUTURE WORK

This work represents an initial exploration into the cost-efficient deployment of NL2SQL systems through strategic integration of SLMs and LLMs. While we have demonstrated significant cost reduction through the hybrid architecture, several avenues for future investigation remain. We plan to conduct a comprehensive analysis of token usage patterns across different query complexity levels to provide more granular insights into the cost-performance trade-offs. Additionally, a detailed study of latency characteristics and end-to-end response times will better quantify the practical deployment implications of our approach. Furthermore, future work will explore additional strategies for privacy preservation during LLM fallback scenarios, including techniques for masking sensitive information in queries sent to external models while maintaining query generation accuracy.

## REFERENCES

[1] B. Li, Y. Luo, C. Chai, G. Li, and N. Tang, "The Dawn of Natural Language to SQL: Are We Fully Ready?," in Proceedings of the VLDB Endowment, VLDB Endowment, 2024, pp. 3318–3331. doi: 10.14778/3681954.3682003.

[2] X. Zhu, Q. Li, L. Cui, and Y. Liu, "Large Language Model Enhanced Text-to-SQL Generation: A Survey," Oct. 2024, [Online]. Available: http://arxiv.org/abs/2410.06011

[3] J. Li et al., "Can LLM Already Serve as A Database Interface? A BIg Bench for Large-Scale Database Grounded Text-to-SQLs." [Online]. Available: https://relational.fit.cvut.cz/

[4] T. Yu et al., "Spider: A Large-Scale Human-Labeled Dataset for Complex and Cross-Domain Semantic Parsing and Text-to-SQL Task," Feb. 2019, [Online]. Available: http://arxiv.org/abs/1809.08887

[5] V. Zhong, C. Xiong, and R. Socher, "Seq2SQL: Generating Structured Queries from Natural Language using Reinforcement Learning," Nov. 2017, [Online]. Available: http://arxiv.org/abs/1709.00103

[6] L. Wang et al., "Plan-and-Solve Prompting: Improving Zero-Shot Chain-of-Thought Reasoning by Large Language Models," May 2023, [Online]. Available: http://arxiv.org/abs/2305.04091

[7] B. Wang et al., "MAC-SQL: A Multi-Agent Collaborative Framework for Text-to-SQL," Mar. 2025, [Online]. Available: http://arxiv.org/abs/2312.11242

[8] Z. Hong et al., "Next-Generation Database Interfaces: A Survey of LLM-based Text-to-SQL," Mar. 2025, [Online]. Available: http://arxiv.org/abs/2406.08426

[9] Y.-T. and C. M. Walker Nick and Peng, "Neural Semantic Parsing with Anonymization for Command Understanding in General-Purpose Service Robots," in RoboCup 2019: Robot World Cup XXIII, Springer International Publishing, 2019, pp. 337–350.

[10] T. Scholak, N. Schucher, and D. Bahdanau, "PICARD: Parsing Incrementally for Constrained Auto-Regressive Decoding from Language Models," Sep. 2021, [Online]. Available: http://arxiv.org/abs/2109.05093

[11] A. Marshan, A. N. Almutairi, A. Ioannou, D. Bell, A. Monaghan, and M. Arzoky, "MedT5SQL: a transformers-based large language model for text-to-SQL conversion in the healthcare domain," Front Big Data, vol. 7, 2024, doi: 10.3389/fdata.2024.1371680.

[12] G. Liu et al., "Solid-SQL: Enhanced Schema-linking based In-context Learning for Robust Text-to-SQL," Dec. 2024, [Online]. Available: http://arxiv.org/abs/2412.12522

[13] OpenAI et al., "GPT-4 Technical Report," Mar. 2024, [Online]. Available: http://arxiv.org/abs/2303.08774

[14] H. Li et al., "OmniSQL: Synthesizing High-quality Text-to-SQL Data at Scale," Jul. 2025, [Online]. Available: http://arxiv.org/abs/2503.02240

[15] Y. Zhu, R. Jiang, B. Li, N. Tang, and Y. Luo, "EllieSQL: Cost-Efficient Text-to-SQL with Complexity-Aware Routing," Aug. 2025, [Online]. Available: http://arxiv.org/abs/2503.22402

[16] S. Zhang et al., "OPT: Open Pre-trained Transformer Language Models," Jun. 2022, [Online]. Available: http://arxiv.org/abs/2205.01068

[17] M. Pourreza and D. Rafiei, "DTS-SQL: Decomposed Text-to-SQL with Small Large Language Models." [Online]. Available: https://anonymous.4open.science/r/

[18] Y. Shen, X. Lin, J. Liu, Z. Huang, S. Wang, and Q. Liu, "SA-SQL: A Schema-Aligned Framework for Text-to-SQL through Large Language Models," in 2024 International Conference on Computational Linguistics and Natural Language Processing (CLNLP), 2024, pp. 71–77. doi: 10.1109/CLNLP64123.2024.00021.

[19] D. Gao et al., "Text-to-SQL Empowered by Large Language Models: A Benchmark Evaluation," Nov. 2023, [Online]. Available: http://arxiv.org/abs/2308.15363

[20] M. Pourreza and D. Rafiei, "DIN-SQL: Decomposed In-Context Learning of Text-to-SQL with Self-Correction." [Online]. Available: https://github.com/MohammadrezaPourreza/Few-shot-NL2SQL-with-prompting

[21] H. Li et al., "CodeS: Towards Building Open-source Language Models for Text-to-SQL," Feb. 2024, [Online]. Available: http://arxiv.org/abs/2402.16347

[22] LangChain AI, "LangGraph documentation," 2024.

[23] A. Grattafiori et al., "The Llama 3 Herd of Models," Nov. 2024, [Online]. Available: http://arxiv.org/abs/2407.21783